\documentclass[11pt,a4paper]{article}
\usepackage[hyperref]{acl2019}
\usepackage{times}
\usepackage{latexsym}
\usepackage{epsfig}
\usepackage{graphicx}

\usepackage{url}

\usepackage{amsmath}
\usepackage{amssymb}

\usepackage[linesnumbered,vlined,ruled,norelsize]{algorithm2e}

\aclfinalcopy 
\def\aclpaperid{24} 

\title{MOROCO: The Moldavian and Romanian Dialectal Corpus}

\author{Andrei M. Butnaru \and {Radu Tudor} Ionescu\\
  Department of Computer Science, University of Bucharest\\
  14 Academiei, Bucharest, Romania\\
  {\tt butnaruandreimadalin@gmail.com}\\
  {\tt raducu.ionescu@gmail.com}
}

\date{}

\begin{document}
\maketitle
\begin{abstract}
In this work, we introduce the {\textbf{Mo}}ldavian and {\textbf{Ro}}manian Dialectal {\textbf{Co}}rpus (MOROCO), which is freely available for download at {https://github.com/butnaruandrei/MOROCO}. The corpus contains 33564 samples of text  (with over 10 million tokens) collected from the news domain. The samples belong to one of the following six topics: culture, finance, politics, science, sports and tech. The data set is divided into 21719 samples for training, 5921 samples for validation and another 5924 samples for testing. For each sample, we provide corresponding dialectal and category labels. This allows us to perform empirical studies on several classification tasks such as $(i)$ binary discrimination of Moldavian versus Romanian text samples, $(ii)$ intra-dialect multi-class categorization by topic and $(iii)$ cross-dialect multi-class categorization by topic. We perform experiments using a shallow approach based on string kernels, as well as a novel deep approach based on character-level convolutional neural networks containing Squeeze-and-Excitation blocks. We also present and analyze the most discriminative features of our best performing model, before and after named entity removal.
\end{abstract}

\setlength{\abovedisplayskip}{3pt}
\setlength{\belowdisplayskip}{3pt}

\section{Introduction}
\label{sec_Introduction}
\vspace*{-0.1cm} 

The high number of evaluation campaigns on spoken or written dialect identification conducted in recent years \cite{Ali-ASRU-2017,Malmasi-VarDial-2016,Rangel-CLEF-2017,Zampieri-VarDial-2017,Zampieri-Vardial-2018} prove that dialect identification is an interesting and challenging natural language processing (NLP) task, actively studied by researchers in nowadays. Due to the recent interest in dialect identification, we introduce the {\textbf{Mo}}ldavian and {\textbf{Ro}}manian Dialectal {\textbf{Co}}rpus (MOROCO), which is composed of 33564 samples of text collected from the news domain. 

\begin{figure}[!t]
\begin{center}
\includegraphics[width=0.78\linewidth]{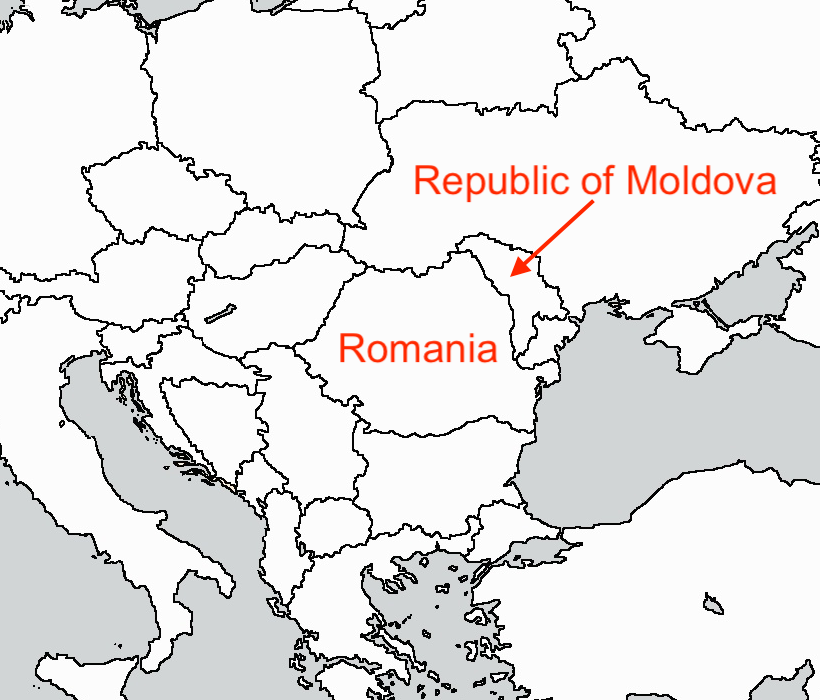}
\end{center}
\vspace*{-0.3cm}
\caption{Map of Romania and the Republic of Moldova.}
\label{fig_ee_map}
\vspace*{-0.4cm}
\end{figure}

Romanian is part of the Balkan-Romance group that evolved from several dialects of Vulgar Latin, which separated from the Western Romance branch of languages from the fifth century \cite{Coteanu-AR-1969}. In order to distinguish Romanian within the Balkan-Romance group in comparative linguistics, it is referred to as \emph{Daco-Romanian}. Along with Daco-Romanian, which is currently spoken in Romania, there are three other dialects in the Balkan-Romance branch, namely Aromanian, Istro-Romanian, and Megleno-Romanian. Moldavian is a subdialect of Daco-Romanian, that is spoken in the Republic of Moldova and in northeastern Romania. The delimitation of the Moldavian dialect, as with all other Romanian dialects, is made primarily by analyzing its phonetic features and only marginally by morphological, syntactical, and lexical characteristics. Although the spoken dialects in Romania and the Republic of Moldova are different, the two countries share the same literary standard \cite{Minahan-R-2013}. Some linguists \cite{Pavel-LR-2008} consider that the border between Romania and the Republic of Moldova (see Figure~\ref{fig_ee_map}) does not correspond to any significant isoglosses to justify a dialectal division. One question that arises in this context is whether we can train a machine to accurately distinguish literary text samples written by people in Romania from literary text samples written by people in the Republic of Moldova. If we can construct such a machine, then what are the discriminative features employed by this machine? Our corpus formed of text samples collected from Romanian and Moldavian news websites, enables us to answer these questions.
Furthermore, MOROCO provides a benchmark for the evaluation of dialect identification methods. To this end, we consider two state-of-the-art methods, string kernels \cite{Butnaru-VarDial-2018,Ionescu-VarDial-2017,Ionescu-EMNLP-2014} and character-level convolutional neural networks (CNNs) \cite{Ali-VarDial-2018,Belinkov-VarDial-2016,Zhang-NIPS-2015}, which obtained the first two places \cite{Ali-VarDial-2018,Butnaru-VarDial-2018} in the Arabic Dialect Identification Shared Task of the 2018 VarDial Evaluation Campaign \cite{Zampieri-Vardial-2018}. We also experiment with a novel CNN architecture inspired the recently introduced Squeeze-and-Excitation (SE) networks \cite{Hu-CVPR-2018}, which exhibit state-of-the-art performance in object recognition from images. To our knowledge, we are the first to introduce Squeeze-and-Excitation networks in the text domain.

As we provide category labels for the collected text samples, we can perform additional experiments on various text categorization by topic tasks. One type of task is intra-dialect multi-class categorization by topic, i.e. the task is to classify the samples written either in the Moldavian dialect or in the Romanian dialect into one of the following six topics: culture, finance, politics, science, sports and tech. Another type of task is cross-dialect multi-class categorization by topic, i.e. the task is to classify the samples written in one dialect, e.g. Romanian, into six topics, using a model trained on samples written in the other dialect, e.g. Moldavian. These experiments are aimed at showing if the considered text categorization methods are robust to the dialect shift between training and testing.

In summary, our contribution is threefold:
\begin{itemize}
 \vspace*{-0.2cm} 
\item We introduce a novel large corpus containing 33564 text samples written in the Moldavian and the Romanian dialects. \vspace*{-0.3cm} 
\item We introduce Squeeze-and-Excitation networks to the text domain. \vspace*{-0.3cm} 
\item We analyze the discriminative features that help the best performing method, string kernels, in $(i)$ distinguishing the Moldavian and the Romanian dialects and in $(ii)$ categorizing the text samples by topic. \vspace*{-0.1cm}
\end{itemize}
 
We organize the remainder of this paper as follows. We discuss related work in Section \ref{sec_Related_Work}.  We describe the MOROCO data set in Section \ref{sec_Corpus}. We present the chosen classification methods in Section \ref{sec_Methods}. We show empirical results in Section \ref{sec_Experiments}, and we provide a discussion on the discriminative features in Section \ref{sec_Discussion}. Finally, we draw our conclusion in Section \ref{sec_Conclusion}.

\vspace*{-0.1cm} 
\section{Related Work}
\label{sec_Related_Work}
\vspace*{-0.1cm} 

There are several corpora available for dialect identification \cite{Ali-INTERSPEECH-2016,Alsarsour-LREC-2018,Bouamor-LREC-2018,Francom-LREC-2014,Johannessen-NODALIDA-2009,Kumar-LREC-2018,Samardzic-LREC-2016,Tan-BUCC-2014,Zaidan-ACL-2011}. Most of these corpora have been proposed for languages that are widely spread across the globe, e.g. Arabic \cite{Ali-INTERSPEECH-2016,Alsarsour-LREC-2018,Bouamor-LREC-2018}, Spanish \cite{Francom-LREC-2014}, Indian \cite{Kumar-LREC-2018} or German \cite{Samardzic-LREC-2016}. Among these, Arabic is the most popular, with a number of four data sets \cite{Ali-INTERSPEECH-2016,Alsarsour-LREC-2018,Bouamor-LREC-2018,Zaidan-ACL-2011}, if not even more.

\noindent
{\bf Arabic.}
The Arabic Online news Commentary (AOC) \cite{Zaidan-ACL-2011} is the first available dialectal Arabic data set. Although AOC contains 3.1 million comments gathered from Egyptian, Gulf and Levantine news websites, the authors labeled only around $0.05\%$ of the data set through the Amazon Mechanical Turk crowdsourcing platform.
\newcite{Ali-INTERSPEECH-2016} constructed a data set of audio recordings, Automatic Speech Recognition transcripts and phonetic transcripts of Arabic speech collected from the Broadcast News domain. The data set was used in the 2016, 2017 and 2018 VarDial Evaluation Campaigns \cite{Malmasi-VarDial-2016,Zampieri-VarDial-2017,Zampieri-Vardial-2018}.
\newcite{Alsarsour-LREC-2018} collected the Dialectal ARabic Tweets (DART) data set, which contains around 25K manually-annotated tweets. The data set is well-balanced over five main groups of Arabic dialects: Egyptian, Maghrebi, Levantine, Gulf and Iraqi.
\newcite{Bouamor-LREC-2018} presented a large parallel corpus of 25 Arabic city dialects, which was created by translating selected sentences from the travel domain.

\noindent
{\bf Other languages.}
The Nordic Dialect Corpus \cite{Johannessen-NODALIDA-2009} contains about 466K spoken words from Denmark, Faroe Islands, Iceland, Norway and Sweden. The authors transcribed each dialect by the standard official orthography of the corresponding country.
\newcite{Francom-LREC-2014} introduced the ACTIV-ES corpus, which represents a cross-dialectal record of the informal language use of Spanish speakers from Argentina, Mexico and Spain. The data set is composed of 430 TV or movie subtitle files.
The DSL corpus collection \cite{Tan-BUCC-2014} comprises news data from various corpora to emulate the diverse news content across different languages. The collection is comprised of six language variety groups. For each language, the collection contains 18K training sentences, 2K validation sentences and 1K test sentences.
The ArchiMob corpus \cite{Samardzic-LREC-2016} contains manually-annotated transcripts of Swiss German speech collected from four different regions: Basel, Bern, Lucerne and Zurich. The data set was used in the 2017 and 2018 VarDial Evaluation Campaigns \cite{Zampieri-VarDial-2017,Zampieri-Vardial-2018}.
\newcite{Kumar-LREC-2018} constructed a corpus of five Indian dialects consisting of 307K sentences. The samples were collected by scanning, passing through an OCR engine and proofreading printed stories, novels and essays from books, magazines or newspapers.

\noindent
{\bf Romanian.}
To our knowledge, the only empirical study on Romanian dialect identification was conducted by \newcite{Ciobanu-LREC-2016}. In their work, \newcite{Ciobanu-LREC-2016} used only a short list of 108 parallel words in a binary classification task in order to discriminate between Daco-Romanian words versus Aromanian, Istro-Romanian and Megleno-Romanian words. Different from \newcite{Ciobanu-LREC-2016}, we conduct a large scale study on 33K documents that contain a total of about 10 million tokens.

\vspace*{-0.1cm} 
\section{MOROCO}
\label{sec_Corpus}
\vspace*{-0.05cm} 

\begin{figure}[!t]
\begin{center}
\includegraphics[width=1.0\linewidth]{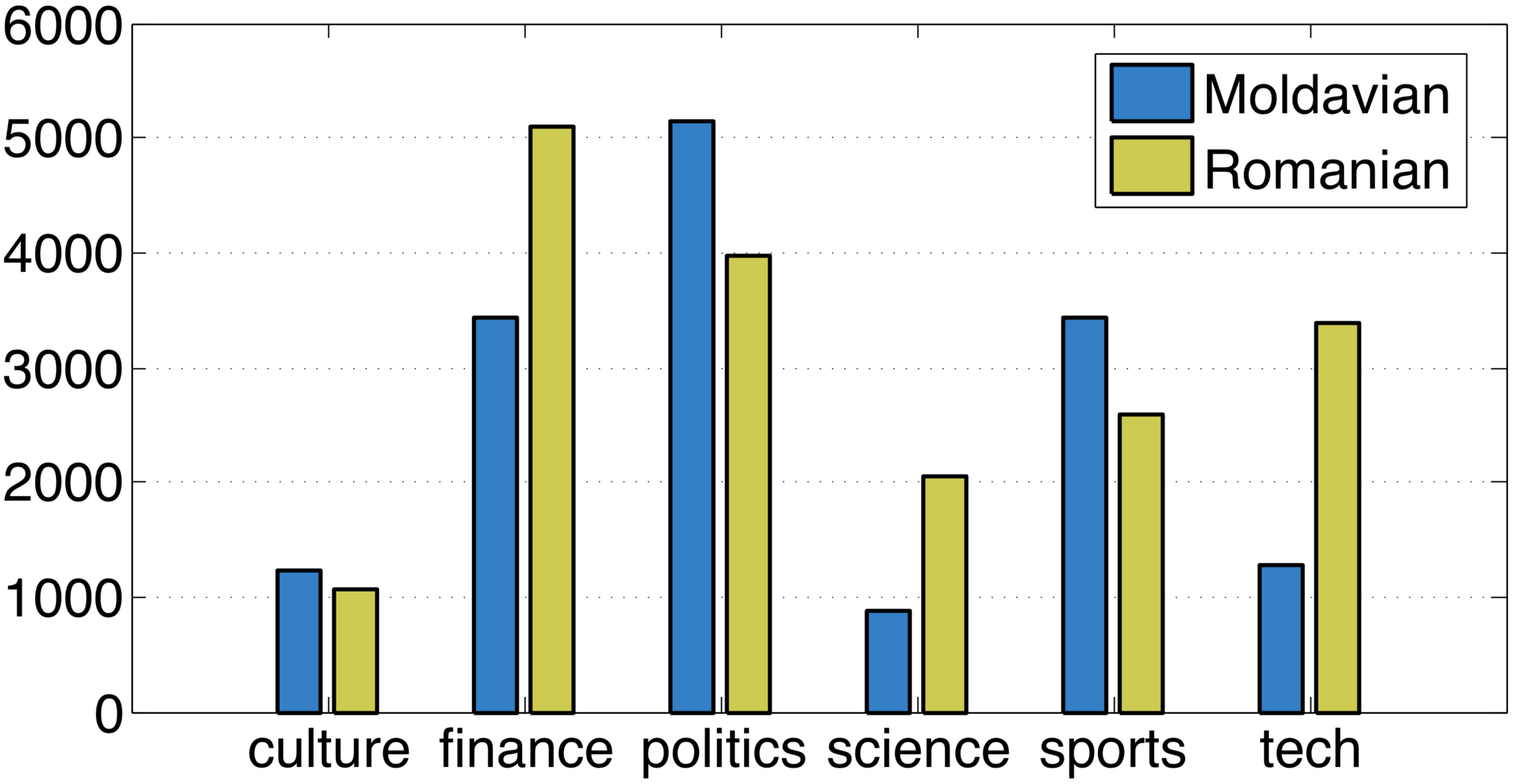}
\end{center}
\vspace*{-0.3cm}
\caption{The distribution of text samples per topic for the Moldavian and the Romanian dialects, respectively. Best viewed in color.}
\label{fig1}
\vspace*{-0.4cm}
\end{figure}

In order to build MOROCO, we collected text samples from the top five most popular news websites in Romania and the Republic of Moldova, respectively. Since news websites in the two countries belong to different Internet domains, the text samples can be automatically labeled with the corresponding dialect. We selected news from six different topics, for which we found at least 2000 text samples in both dialects. For each dialect, we illustrate the distribution of text samples per topic in Figure \ref{fig1}. In both countries, we notice that the most popular topics are finance and politics, while the least popular topics are culture and science. The distribution of topics for the two dialects is mostly similar, but not very well-balanced. For instance, the number of Moldavian politics samples (5154) is about six times higher than the number of Moldavian science samples (877). However, MOROCO is well-balanced when it comes to the distribution of samples per dialect, since we were able to collect 15403 Moldavian text samples and 18161 Romanian text samples. 

It is important to note that, in order to obtain the text samples, we removed all HTML tags and replaced consecutive space characters with a single space character. We further processed the samples in order to eliminate named entities. Previous research \cite{AbuJbara-BEA8-2013,Nicolai-ACL-2014} found that named entities such as country names or cities can provide clues about the native language of English learners. We decided to remove named entities in order to prevent classifiers from taking the decision based on features that are not truly indicative of the dialects or the topics. For example, named entities representing city names in Romania or Moldova can provide clues about the dialect, while named entities representing politicians or football players names can provide clues about the topic. The identified named entities are replaced with the token \emph{\$NE\$}. In the experiments, we present results before and after named entity removal, in order to illustrate the effect of named entities.

\begin{table}[!t]
\setlength\tabcolsep{5.5pt}
\small{
\begin{center}
\begin{tabular}{|l|r|r|}
\hline
Set 						& \#samples						&	\#tokens	\\
\hline
\hline
\vspace{-0.9em}\\
Training				& 21,719 							& 6,705,334 \\
Validation				& 5,921								& 1,826,818 \\
Test						& 5,924							& 1,850,977 \\
\hline
Total						& 33,564							& 10,383,129 \\
\hline
\end{tabular}
\end{center}
}
\vspace*{-0.1cm}
\caption{The number of samples (\#samples) and the number of tokens (\#tokens) contained in the training, validation and test sets included in our corpus.}
\label{tab_stats}
\vspace*{-0.2cm}
\end{table}

In order to allow proper comparison in future research, we divided MOROCO into a training, a validation and a test set. We used stratified sampling in order to produce a split that preserves the distribution of dialects and topics across all subsets. Table \ref{tab_stats} shows some statistics of the number of samples as well as the number of tokens in each subset.  We note that the entire corpus contains 33564 samples with more than 10 million tokens in total. On average, there are about 309 tokens per sample.

Since we provide both dialectal and category labels for each sample, we can perform several tasks on MOROCO:
\begin{itemize}
\vspace*{-0.2cm} 
\item Binary classification by dialect -- the task is to discriminate between the Moldavian and the Romanian dialects.\vspace*{-0.3cm} 
\item Moldavian (MD) intra-dialect multi-class categorization by topic -- the task is to classify the samples written in the Moldavian dialect into six topics.\vspace*{-0.3cm}
\item Romanian (RO) intra-dialect multi-class categorization by topic -- the task is to classify the samples written in the Romanian dialect into six topics.\vspace*{-0.3cm}
\item MD$\rightarrow$RO cross-dialect multi-class categorization by topic -- the task is to classify the samples written in the Romanian dialect into six topics, using a model trained on samples written in the Moldavian dialect.\vspace*{-0.3cm} 
\item RO$\rightarrow$MD cross-dialect multi-class categorization by topic -- the task is to classify the samples written in the Moldavian dialect into six topics, using a model trained on samples written in the Romanian dialect. \vspace*{-0.1cm} 
\end{itemize}

\section{Methods}
\label{sec_Methods}
\vspace*{-0.05cm} 

\noindent
{\bf String kernels.}
Kernel functions \cite{taylor-Cristianini-cup-2004} capture the intuitive notion of similarity between objects in a specific domain. For example, in text mining, string kernels can be used to measure the pairwise similarity between text samples, simply based on character n-grams. Various string kernel functions have been proposed to date \cite{Ionescu-EMNLP-2014,LodhiSSCW02,taylor-Cristianini-cup-2004}. Recently, the presence bits string kernel and the histogram intersection kernel obtained state-of-the-art results in a broad range of text classification tasks such as dialect identification \cite{Butnaru-VarDial-2018,Ionescu-VarDial-2017,Ionescu-VarDial-2016}, native language identification \cite{Ionescu-COLI-2016,Ionescu-BEA-2017}, sentiment analysis \cite{franco-EACL-2017,Ionescu-EMNLP-2018,marius-KES-2017} and automatic essay scoring \cite{Cozma-ACL-2018}. In this paper, we opt for the presence bits string kernel, which allows us to derive the primal weights and analyze the most discriminative features, as explained by \newcite{Ionescu-COLI-2016}. For two strings over an alphabet $\Sigma$, $x,y \in \Sigma^*$, the presence bits string kernel is formally defined as:
\begin{equation*}
\begin{split}
k^{0/1}_n(x,y)=\sum\limits_{s \in \Sigma^n}\mbox{in}_s(x) \cdot \mbox{in}_s(y),
\end{split}
\end{equation*}
where $\mbox{in}_s(x)$ is $1$ if string $s$ occurs as a substring in $x$, and $0$ otherwise. In our empirical study, we experiment with character n-grams in a range, and employ the Kernel Ridge Regression (KRR) binary classifier. During training, KRR finds the vector of weights that has both small empirical error and small norm in the Reproducing Kernel Hilbert Space generated by the kernel function. The ratio between the empirical error and the norm of the weight vector is controlled through the regularization parameter $\lambda$.

\noindent
{\bf Character-level CNN.}
Convolutional networks \cite{lecun-bottou-ieee-1998,Hinton-NIPS-2012} have been employed for solving many NLP tasks such as part-of-speech tagging \cite{Santos-ICML-2014}, text categorization \cite{Johnson-NAACL-2015,Kim-EMNLP-2014,Zhang-NIPS-2015}, dialect identification \cite{Ali-VarDial-2018,Belinkov-VarDial-2016}, machine translation \cite{Gehring-ACL-2017} and language modeling \cite{Dauphin-ICML-2017,Kim-AAAI-2016}. Many CNN-based methods rely on words, the primary reason for this being the aid given by word embeddings \cite{Mikolov-NIPS-2013,Pennington-EMNLP-2014} and their ability to learn semantic and syntactic latent features. Trying to eliminate the pre-trained word embeddings from the pipeline, some researchers have decided to build end-to-end models using characters as input, in order to solve text classification \cite{Zhang-NIPS-2015,Belinkov-VarDial-2016} or language modeling tasks \cite{Kim-AAAI-2016}. At the character-level, the model can learn unusual character sequences such as misspellings or take advantage of unseen words during test time. This appears to be particularly helpful in dialect identification, since some state-of-the-art dialect identification methods \cite{Butnaru-VarDial-2018,Ionescu-VarDial-2017} use character n-grams as features.

\begin{figure}[!t]
\begin{center}
\includegraphics[width=1.0\linewidth]{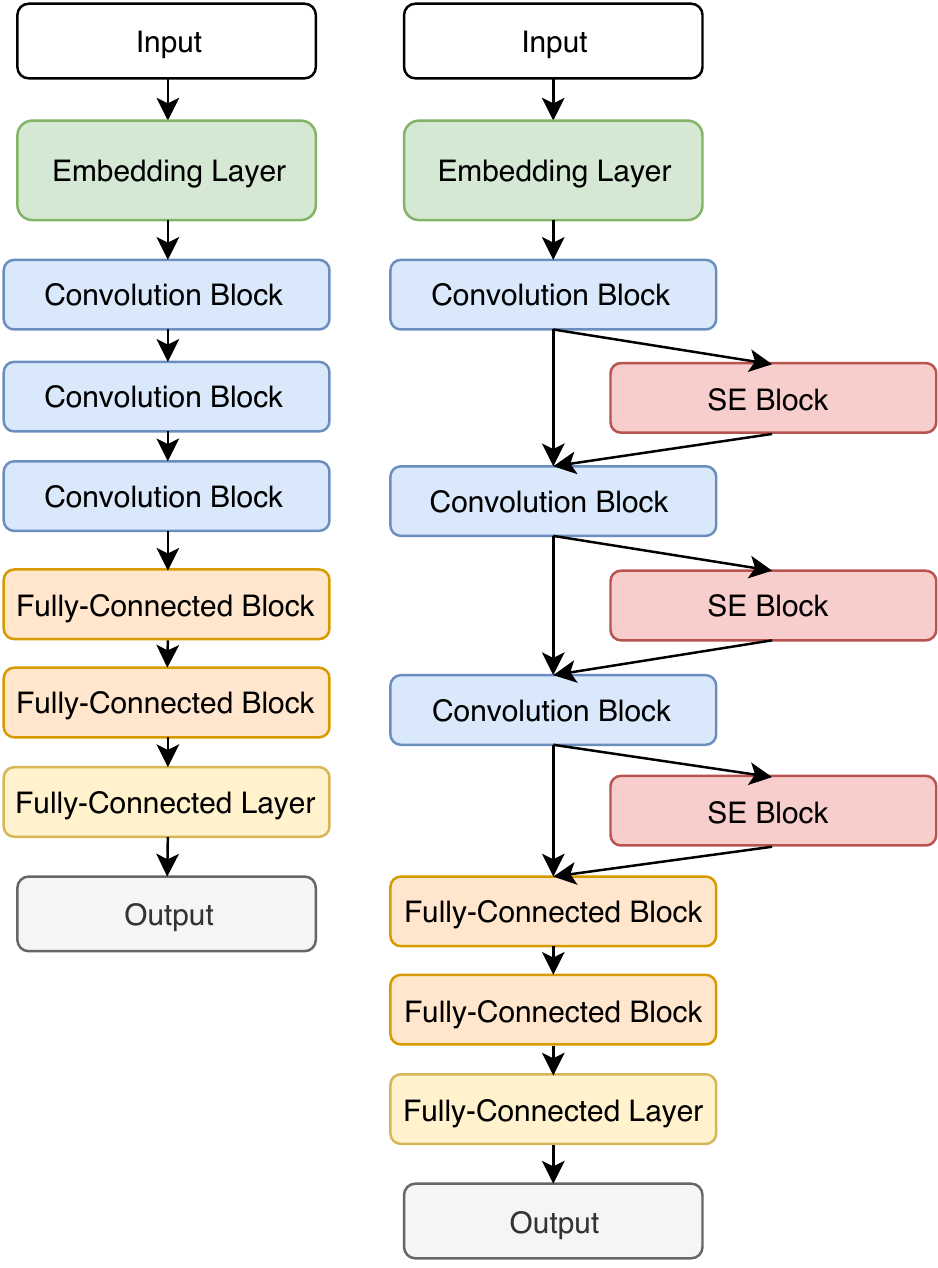}
\end{center}
\vspace*{-0.2cm}
\caption{\textbf{Left:} The architecture of the baseline character-level CNN composed of seven blocks. \textbf{Right:} The modified CNN architecture, which includes a Squeeze-and-Excitation block after each convolutional block.}
\label{fig2}
\vspace*{-0.2cm}
\end{figure}

In this paper, we draw our inspiration from \newcite{Zhang-NIPS-2015} in order to design a lightweight character-level CNN architecture for dialect identification. One way proposed by \newcite{Zhang-NIPS-2015} to represent characters in a character-level CNN is to map every character from an alphabet of size $t$ to a discrete value using a $1$-of-$t$ encoding. For example, having the alphabet $\Sigma = \{a,b,c\}$, the encoding for the character $a$ is $1$, for $b$ is 2, and for $c$ is $3$. Each character from the input text is encoded, and only a fixed size $l$ of the input is kept. In our case, we keep the first $l=5000$ characters, zero-padding the documents that are under length. We compose an alphabet of $105$ characters that includes uppercase and lowercase characters, Moldavian and Romanian diacritics (such as \u{a}, \^{a}, \^{i}, \c{s} and \c{t}), digits, and $33$ other symbol characters. Characters that do not appear in the alphabet are encoded as a blank character.

As illustrated in the left-hand side of Figure \ref{fig2}, our architecture is seven blocks deep, containing one embedding layer, three convolutional and max-pooling blocks, and three fully-connected blocks. The first two convolutional layers are based on one-dimensional filters of size $7$, the third one being based on one-dimensional filters of size $3$. A thresholded Rectified Linear Units (ReLU) activation function \cite{Nair-ICML-2010} follows each convolutional layer. The max-pooling layers are based on one-dimensional filters of size $3$ with stride $3$. After the third convolutional block, the activation maps pass through two fully-connected blocks having thresholded ReLU activations. Each of these two fully-connected blocks is followed by a dropout layer with the dropout rate of $0.5$. The last fully-connected layer is followed by softmax, which provides the final output. All convolutional layers have $128$ filters, and the threshold used for the thresholded ReLU is $10^{-6}$. The network is trained with the Adam optimizer \cite{Kingma-ICLR-2015} using categorical cross-entropy as loss function. 

\begin{figure*}[!t]
\begin{center}
\includegraphics[width=0.92\linewidth]{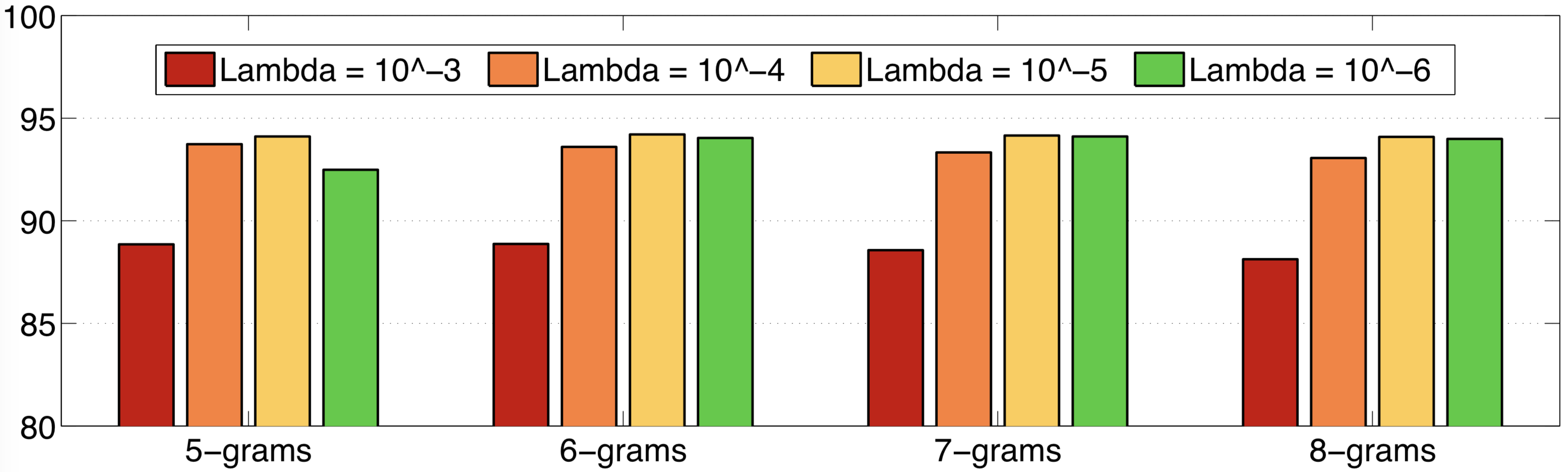}
\end{center}
\vspace*{-0.2cm}
\caption{Dialect classification results on the validation set provided by the KRR classifier based on the presence bits string kernel with n-grams in the range 5-8. Results are reported for various $\lambda$ values from $10^{-3}$ to $10^{-6}$. Best viewed in color.}
\label{fig3}
\end{figure*}

\noindent
{\bf Squeeze-and-Excitation Networks.}
\newcite{Hu-CVPR-2018} argued that the convolutional filters close to the input layer are not aware of the global appearance of the objects in the input image, as they operate at the local level. To alleviate this problem, \newcite{Hu-CVPR-2018} proposed to insert Squeeze-and-Excitation blocks after each convolutional block that is closer to the network's input. The SE blocks are formed of two layers, \emph{squeeze} and \emph{excitation}. The activation maps of a given convolutional block are first passed through the squeeze layer, which aggregates the activation maps across the spatial dimension in order to produce a channel descriptor. This layer can be implemented through a global average pooling operation. In our case, the size of the output after the squeeze operation is $1 \times 128$, since our convolutional layers are one-dimensional and each layer contains $d=128$ filters. The resulting channel descriptor enables information from the global receptive field of the network to be leveraged by the layers near the network's input. The squeeze layer is followed by an excitation layer based on a self-gating mechanism, which aims to capture channel-wise dependencies. The self-gating mechanism is implemented through two fully-connected layers, the first being followed by ReLU activations and the second being followed by sigmoid activations, respectively. The first fully-connected layer acts as a bottleneck layer, reducing the input dimension (given by the number of filters $d$) with a reduction ratio $r$. This is achieved by assigning $d/r$ units to the bottleneck layer. The second fully-connected layer increases the size of the output back to $1 \times 128$. Finally, the activation maps of the preceding convolutional block are then reweighted (using the $1 \times 128$ outputs provided by the excitation layer as weights) to generate the output of the SE block, which can then be fed directly into subsequent layers. Thus, SE blocks are just alternative pathways designed to recalibrate channel-wise feature responses by explicitly modeling interdependencies between channels. We insert SE blocks after each convolutional block, as illustrated in the right-hand side of Figure \ref{fig2}.

\vspace*{-0.1cm} 
\section{Experiments}
\label{sec_Experiments}

\begin{table*}[!t]
\setlength\tabcolsep{3.0pt}
\small{
\begin{center}
\begin{tabular}{|c|l|ccc|ccc|}
\hline
Task									& Method 						& \multicolumn{3}{|c|}{Validation}									& \multicolumn{3}{|c|}{Test}\\		
\cline{3-8}
										&									&  accuracy  		& weighted $F_1$		& macro $F_1$		&  accuracy  		& weighted $F_1$		& macro $F_1$\\
\hline
\hline
Binary								& KRR + $k^{0/1}_6$		& $94.21$ 		& $94.20$ 				& $94.15$				& $94.13$			& $94.11$					& $94.06$\\
classification						& CNN							& $93.00$ 		& $93.00$ 				& $92.95$			& $92.75$		& $92.76$				& $92.71$\\
by dialect							& CNN + SE					& $93.02$ 		& $93.01$ 				& $92.95$			& $92.99$		& $92.98$				& $92.93$\\
\hline
MD									& KRR + $k^{0/1}_6$		& $92.49$ 		& $92.46$ 				& $90.45$			& $92.68$		& $92.63$				& $90.57$\\
categorization					& CNN							& $85.42$ 		& $85.25$ 				& $79.28$			& $83.92$		& $83.73$				& $76.82$\\
(by topic)							& CNN + SE					& $86.23$ 		& $85.97$ 				& $80.51$				& $84.39$		& $84.01$					& $77.85$\\
\hline
MD$\rightarrow$RO			& KRR + $k^{0/1}_6$		& $92.49$ 		& $92.46$ 				& $90.45$			& $68.21$			& $67.59$				& $67.59$\\
categorization					& CNN							& $85.42$ 		& $85.25$ 				& $79.28$			& $55.04$		& $56.27$				& $53.67$\\
(by topic)							& CNN + SE					& $86.23$ 		& $85.97$ 				& $80.51$				& $56.31$			& $57.01$					& $53.85$\\
\hline
RO									& KRR + $k^{0/1}_6$		& $74.38$ 		& $74.49$ 				& $77.68$			& $74.98$		& $75.11$					& $78.76$\\
categorization					& CNN							& $68.04$ 		& $67.10$ 				& $63.84$			& $68.14$		& $67.43$				& $64.98$\\
(by topic)							& CNN + SE					& $68.76$ 		& $68.67$ 				& $67.74$			& $69.04$		& $69.07$				& $68.77$\\
\hline
RO$\rightarrow$MD			& KRR + $k^{0/1}_6$		& $74.38$ 		& $74.49$ 				& $77.68$			& $82.31$			& $82.17$					& $75.47$\\
categorization					& CNN							& $68.04$ 		& $67.10$ 				& $63.84$			& $72.49$		& $71.60$					& $62.70$\\
(by topic)							& CNN + SE					& $68.76$ 		& $68.67$ 				& $67.74$			& $74.84$		& $74.87$				& $67.42$\\
\hline
\end{tabular}
\end{center}
}
\vspace*{-0.1cm}
\caption{Accuracy rates, weighted $F_1$ scores and macro-averaged $F_1$-scores (in $\%$) for the five evaluation tasks: binary classification by dialect, Moldavian intra-dialect 6-way categorization (by topic), MD$\rightarrow$RO cross-dialect 6-way categorization, Romanian (RO) intra-dialect 6-way categorization, and RO$\rightarrow$MD cross-dialect 6-way categorization. Results are reported for three baseline models: KRR based on the presence bits string kernel (KRR+$k^{0/1}_6$), convolutional neural networks (CNN), and Squeeze-and-Excitation convolutional neural networks (CNN+SE).}
\label{tab_results}
\vspace*{-0.2cm}
\end{table*}

\noindent
{\bf Parameter tuning.}
In order to tune the parameters of each model, we used the MOROCO validation set. We first carried out a set of preliminary dialect classification experiments to determine the optimal choice of n-grams length for the presence bits string kernel and the regularization parameter $\lambda$ of the KRR classifier. We present results for these preliminary experiments in Figure~\ref{fig3}. We notice that both $\lambda=10^{-4}$ and $\lambda=10^{-5}$ are good regularization choices, with $\lambda=10^{-5}$ being slightly better for all n-grams lengths between 5 and 8. Although 6-grams, 7-grams and 8-grams attain almost equally good results, the best choice according to the validation results is to use 6-grams. Therefore, in the subsequent experiments, we employ the presence bits string kernel based on n-grams of length 6 and KRR with $\lambda=10^{-5}$.

For the baseline CNN, we set the learning rate to $5 \cdot 10^{-4}$ and use mini-batches of $128$ samples during training. We use the same parameters for the SE network. Both deep networks are trained for $50$ epochs. For the SE blocks, we set the reduction ratio to $r = 64$, which results in a bottleneck layer with two neurons. We also tried lower reduction ratios, e.g. 32 and 16, but we obtained lower performance for these values.

\noindent
{\bf Results.}
In Table~\ref{tab_results} we present the accuracy, the weighted $F_1$-scores and the macro-averaged $F_1$-scores obtained by the three classification models (string kernels, CNN and SE networks) for all the classification tasks, on the validation set as well as the test set. Regarding the binary classification by dialect task, we notice that all models attain good results, above $90\%$. SE blocks bring only minor improvements over the baseline CNN. Our deep models, CNN and CNN+SE, attain results around $93\%$, while the string kernels obtain results above $94\%$. We thus conclude that written text samples from the Moldavian and the Romanian dialects can be accurately discriminated by both shallow and deep learning models. This answers our first question from Section~\ref{sec_Introduction}.

Regarding the Moldavian intra-dialect 6-way categorization (by topic) task, we notice that string kernels perform quite well in comparison with the CNN and the CNN+SE models. In terms of the macro-averaged $F_1$ scores, SE blocks bring improvements higher than $1\%$ over the baseline CNN. In the MD$\rightarrow$RO cross-dialect 6-way categorization task, our models attain the lowest performance on the Romanian test set. We would like to note that in both cross-dialect settings, we use the validation set from the same dialect as the training set, in order to prevent any use of information about the test dialect during training. In other words, the settings are intra-dialect with respect to the validation set and cross-dialect with respect to the test set. The Romanian intra-dialect 6-way categorization task seems to be much more difficult than the Moldavian intra-dialect categorization task, since all models obtain scores that are roughly $20\%$ lower. In terms of the macro-averaged $F_1$ scores, SE blocks bring improvements of around $4\%$ over the baseline CNN. However, the results of CNN+SE are still much under those of the presence bits string kernel. Regarding the RO$\rightarrow$MD cross-dialect 6-way categorization task, we find that the models learned on the Romanian training set obtain better results on the Moldavian (cross-dialect) test set than on the Romanian (intra-dialect) test set. Once again, this provides additional evidence that the 6-way categorization by topic task is more difficult for Romanian than for Moldavian. In all the intra-dialect or cross-dialect 6-way categorization tasks, we observe a high performance gap between deep and shallow models. These results are consistent with the recent reports of the VarDial evaluation campaigns \cite{Malmasi-VarDial-2016,Zampieri-VarDial-2017,Zampieri-Vardial-2018}, which point out that shallow approaches such as string kernels \cite{Butnaru-VarDial-2018,Ionescu-VarDial-2017} surpass deep models in dialect and similar language discrimination tasks. Although deep models obtain generally lower results, our proposal of integrating Squeeze-and-Excitation blocks seems to be a steady step towards improving CNN models for language identification, as SE blocks improve performance across all the experiments presented in Table~\ref{tab_results}, and, in some cases, the performance gains are considerable.

\vspace*{-0.1cm} 
\section{Discussion}
\label{sec_Discussion}
\vspace*{-0.05cm} 

\begin{table}[!t]
\setlength\tabcolsep{2.0pt}
\small{
\begin{center}
\begin{tabular}{|c|c|ccc|}
\hline
Task													& NER 				& \multicolumn{3}{|c|}{Test}\\		
\cline{3-5}
														&						&  accuracy  		& weighted $F_1$		& macro $F_1$\\
\hline
\hline
Classification									& No					& $95.61$			& $95.60$				& $95.56$\\
by dialect											& Yes				& $94.13$			& $94.11$					& $94.06$\\
\hline
MD													& No					& $93.23$		& $93.19$					& $91.36$\\
categorization									& Yes				& $92.68$		& $92.63$				& $90.57$\\
\hline
MD$\rightarrow$RO							& No					& $68.80$		& $68.23$				& $68.49$\\
categorization									& Yes				& $68.21$			& $67.59$				& $67.59$\\
\hline
RO													& No					& $76.07$		& $76.19$					& $80.10$\\
categorization									& Yes				& $74.98$		& $75.11$					& $78.76$\\
\hline
RO$\rightarrow$MD							& No					& $82.57$		& $82.46$				& $76.00$\\
categorization									& Yes				& $82.31$			& $82.17$					& $75.47$\\
\hline
\end{tabular}
\end{center}
}
\vspace*{-0.1cm}
\caption{Accuracy rates, weighted $F_1$ scores and macro-averaged $F_1$-scores (in $\%$) of the KRR based on the presence bits string kernel for the five evaluation tasks, before and after named entity removal (NER).}
\label{tab_results_NER}
\end{table}

\begin{table}[!t]
\setlength\tabcolsep{2.4pt}
\small{
\begin{center}
\begin{tabular}{|c|c|c|c|c|}
\hline
NER							& \multicolumn{2}{|c|}{Top 6-grams for MD}				& \multicolumn{2}{|c|}{Top 6-grams for RO}\\
\cline{2-5}
								& original						& translation							& original							& translation\\
\hline
\hline
								& {\bf[P\u{a}m\^{i}nt]} 	& Earth									& {\bf[Rom\^{a}ni]}a 			& Romania\\
								& {\bf[Moldov]}a			& Moldova							& n{\bf[ews.ro]} 				& a website\\
No							& {\bf[c\^{i}teva]}			& some									& {\bf[P\u{a}m\^{a}nt]} 		& Earth\\
								& M{\bf[oldova]} 			& Moldova							& Nicu{\bf[lescu ]} 			& family name\\
								& cuv{\bf[\^{i}ntul ]} 		& the word 							& {\bf[Bucure]}\c{s}ti 		& Bucharest\\
\hline
								& {\bf[ s\^{i}nt ]}			& am / are								& {\bf[ rom\^{a}n]}esc 		& Romanian\\
								& {\bf[ c\^{i}nd ]}			& when									& {\bf[ jude\c{t}]} 				& county\\
Yes							& {\bf[dec\^{i}t ]}			& than									& {\bf[ c\^{a}nd ]}				& when	\\
								& t{\bf[enisme]}n 			& tennis player						& {\bf[ firme]}					& companies\\
								& {\bf[ p\^{i}n\u{a} ]}		& until									& {\bf[ vorbi]} 					& talk\\
\hline
\end{tabular}
\end{center}
}
\vspace*{-0.1cm}
\caption{Examples of n-grams from the Moldavian and the Romanian dialects, that are weighted as more discriminative by the KRR based on the presence bits string kernel, before and after named entity removal (NER). The n-grams are placed between squared brackets and highlighted in bold. The n-grams are posed inside words and translated to English.}
\label{tab_features_dialect}
\vspace*{-0.1cm}
\end{table}

\begin{table*}[!t]
\setlength\tabcolsep{2.2pt}
\small{
\begin{center}
\begin{tabular}{|c|c|c|c|c|c|c|}
\hline
NER							& \multicolumn{2}{|c|}{Top 6-grams for culture}		& \multicolumn{2}{|c|}{Top 6-grams for finance}		& \multicolumn{2}{|c|}{Top 6-grams for politics}		\\
\cline{2-7}
								& original						& translation							& original							& translation						& original							& translation\\
\hline
\hline
								& {\bf[teatru]} 				& theater								& {\bf[econom]}ie 				& economy						& {\bf[. PSD ]} 			& Social-Democrat Party\\
								& {\bf[ scen\u{a}]}			& scene								& {\bf[achita]}t 					& payed							& {\bf[parlam]}ent 			& parliament\\
No							& {\bf[Eurovi]}sion 		& Eurovision contest				& {\bf[tranza]}c\c{t}ie 		& transaction					& Liviu D{\bf[ragnea]} 		& ex-leader of PSD\\
								& {\bf[scriit]}or				& writer									& di{\bf[n Mold]}ova 			& of Moldova					& Igor{\bf[ Dodon]} 			& president of Moldova\\
								& Euro{\bf[vision]} 		& Eurovision	contest				& Un{\bf[iCredi]}t 				& UniCredit Bank				& Dacian {\bf[Ciolo\c{s}]} 	& ex-prime minster of Romania\\
\hline
								& {\bf[muzic\u{a}]} 		& music								& {\bf[ b\u{a}nci]} 				& banks							& {\bf[politi]}ca 				& the politics\\
								& {\bf[ pies\u{a}]}			& piece									& {\bf[monede]} 				& currencies						& {\bf[pre\c{s}ed]}inte 		& president\\
Yes							& {\bf[artist]}				& artist									& {\bf[afacer]}i 					& business						& {\bf[primar]} 					& mayor\\
								& {\bf[actoru]}l 				& the actor							& {\bf[export]}uri 				& exports							& p{\bf[artidu]}l 				& the party\\
								& s{\bf[pectac]}ol 			& show									& p{\bf[roduse]} 				& products						& {\bf[democr]}a\c{t}ie 		& democracy\\
\hline
\hline
								& \multicolumn{2}{|c|}{Top 6-grams for science}		& \multicolumn{2}{|c|}{Top 6-grams for sports}			& \multicolumn{2}{|c|}{Top 6-grams for tech}\\
\hline
\hline
								& {\bf[studiu]} 				& study									& {\bf[Simona]} Halep 		& a tennis player				& {\bf[Intern]}et 				& Internet\\
								& \c{s}{\bf[tiin\c{t}\u{a}]}	& science							& {\bf[campio]}n 				& champion						& Fac{\bf[cebook]} 			& Facebook\\
No							& {\bf[ NASA ]} 				& NASA									& Simona{\bf[ Halep]}  		& a tennis player				& Mol{\bf[dtelec]}om 	& telecom operator in Moldova\\
								& Max {\bf[Planck]}		& Max Planck						& o{\bf[limpic]} 				& Olympic							& com{\bf[unica\c{t}]}ii 		& communications\\
								& {\bf[P\u{a}m\^{i}nt]} 	& Earth									& {\bf[echipe]} 					& teams							& {\bf[ telev]}iziune 			& television\\
\hline
								& {\bf[cercet]}are 			& research							& {\bf[fotbal]} 					& football							& {\bf[ma\c{s}ini]} 			& cars\\
								& {\bf[astron]}omie 		& astronomy							& {\bf[meciul]} 					& the match						& {\bf[utiliz]}ator 				& user\\
Yes							& {\bf[planet]}a				& the planet							& {\bf[juc\u{a}to]}r 			& player							& t{\bf[elefon]} 					& telephone\\
								& {\bf[univer]}sitatea 	& the university						& {\bf[antren]}orul 			& the coach						& {\bf[ compa]}nie 			& company\\
								& {\bf[teorie]} 				& theory								& {\bf[clubul]} 					& the club							& {\bf[tehnol]}ogie 			& technology\\
\hline
\end{tabular}
\end{center}
}
\vspace*{-0.1cm}
\caption{Examples of n-grams from the six different categories in MOROCO, that are weighted as more discriminative by the KRR based on the presence bits string kernel, before and after named entity removal (NER). The n-grams are placed between squared brackets and highlighted in bold. The n-grams are posed inside words and translated to English.}
\label{tab_features_category}
\vspace*{-0.2cm}
\end{table*}

\noindent
{\bf Named entity removal.}
In Table~\ref{tab_results_NER}, we presents comparative results before and after named entity removal (NER). We selected only the KRR based on the presence bits string kernel for this comparative study, since it provides the best performance among the considered baselines. The experiment reveals that named entities can artificially raise the performance by more than $1\%$ in some cases, which is consistent with observations in previous works \cite{AbuJbara-BEA8-2013,Nicolai-ACL-2014}. 

\noindent
{\bf Discriminative features.}
In order to understand why the KRR based on the presence bits string kernel works so well in discriminating the Moldavian and the Romanian dialects, we conduct an analysis of some of the most discriminative features (n-grams), which are listed in Table~\ref{tab_features_dialect}. When named entities are left in place, the classifier chooses the country names (Moldova and Romania) or the capital city of Romania (Bucharest) as discriminative features. When named entities are removed, it seems that Moldavian words that contain the letter `\^{i}' inside, e.g. `c\^{i}nd', are discriminative, since in Romanian, the letter `\^{i}' is only used at the beginning of a word (inside Romanian words, the same sound is denoted by `\^{a}', e.g. `c\^{a}nd'). While Moldavian writers prefer to use `tenismen' to denote `tennis player', Romanians prefer to use `juc\u{a}tor de tenis' for the same concept. Although both terms, `tenismen' and `juc\u{a}tor de tenis', are understood in Romania and the Republic of Moldova, our analysis reveals that preference for one term or the other is not the same.

In a similar manner, we look at examples of features weighted as discriminative by the KRR based on the presence bits string kernel for categorization by topic. Table~\ref{tab_features_category} lists discriminative n-grams for all the six categories inside MOROCO, before and after NER. When named entities are left in place, we notice that the KRR classifier selects some interesting named entities as discriminative. For example, news in the politics domain make a lot of references to politicians such as Liviu Dragnea (the ex-leader of the Social-Democrat Party in Romania), Igor Dodon (the current president of Moldova) or Dacian Ciolo\c{s} (an ex-prime minster of Romania). News that mention NASA (the National Aeronautics and Space Administration) or the Max Planck institute are likely to be classified in the science domain by KRR+$k^{0/1}_6$. After Simona Halep reached the first place in the Women's Tennis Association (WTA) ranking, a lot of sports news that report on her performances started to appear, which determines the classifier to choose `Simona' or ` Halep' as discriminative n-grams. References to the Internet or the Facebook social network indicate that the respective news are from the tech domain, according to our classifier. When named entities are removed, KRR seems to choose plausible words for each category. For instance, it relies on n-grams such as `muzic\u{a}' or `artist' to classify a news sample into the culture domain, or on n-grams such as `campion' or `fotbal' to classify a news sample into the sports domain.

\begin{table}[!t]
\setlength\tabcolsep{2.0pt}
\small{
\begin{center}
\begin{tabular}{|c|c|c|c|}
\hline
Corpus								& \#dialects 		& \#tokens 				&	Accuracy\\		
										&						&	per sample			&\\

\hline
\hline
Romanian	(ours)				& 2					& 309.3					& $94.13$	\\
\hline
Arabic								& 5					& 	22.6						& $76.27$\\
\hline
German								& 4					&	7.9						& $66.36$\\
\hline
\end{tabular}
\end{center}
}
\vspace*{-0.1cm}
\caption{Accuracy rates (in $\%$) of the KRR based on string kernels for Romanian dialect identification versus Arabic \cite{Ali-INTERSPEECH-2016} and German \cite{Samardzic-LREC-2016} dialect identification, respectively. The results for the Arabic and German dialect identification tasks are taken from our previous work \cite{Ionescu-VarDial-2017}. For each corpus, we include the number of dialects (\#dialects) and the average number of tokens in each sample (\#tokens per sample).}
\label{tab_results_other_dialects}
\end{table}

\noindent
{\bf Difficulty with respect to other dialects.}
In our previous work \cite{Ionescu-VarDial-2017}, we have applied the KRR based on string kernels for Arabic dialect identification and German dialect identification. In the case of Arabic, we have reached performance levels of around $76\%$ for discriminating between five dialects. In the same time, we have reached performance levels of around $66\%$ for discriminating between four German dialects. As shown in Table~\ref{tab_results_other_dialects}, it seems to be much easier to discriminate between Romanian dialects, as the accuracy is near $94\%$. However, there are some important differences between these tasks. First of all, the random chance baseline is much high for our binary classification task, as we only have to choose between two dialects: Moldavian or Romanian. Second of all, the number of tokens per sample is much higher for the samples in our corpus compared to the samples provided in the Arabic \cite{Ali-INTERSPEECH-2016} or the German \cite{Samardzic-LREC-2016} corpora. Before drawing the conclusion that Romanian dialects are easier to discriminate than other dialects, we have to make sure that the experiments are conducted in similar conditions. We leave this discussion for future work.

\vspace*{-0.15cm} 
\section{Conclusion}
\label{sec_Conclusion}
\vspace*{-0.15cm} 

In this paper, we presented a novel and large corpus of Moldavian and Romanian dialects. We also introduced Squeeze-and-Excitation networks to the NLP domain, performing comparative experiments using shallow and deep state-of-the-art baselines. We would like to stress out that the methods presented in this paper are only provided as baselines in order to enable comparisons in future work. Our intention was not that of providing top accuracy rates on the MOROCO corpus. In this context, we acknowledge that better accuracy rates can be obtained by combining string kernels using a range of n-grams, as we have already shown for other dialects and tasks in our previous works \cite{Butnaru-VarDial-2018,Cozma-ACL-2018,Ionescu-VarDial-2017,Ionescu-EMNLP-2018}. Another option for improving performance is to combine string kernels and neural networks into an ensemble model. We leave these ideas for future exploration.

Although Romanian and Moldavian are supposed to be hard to discriminate, since Romania and the Republic of Moldova share the same literary standard \cite{Minahan-R-2013}, the empirical results seem to point in the other direction, to our surprise. However, we should note that the high accuracy rates attained by the proposed classifiers can be explained through a combination of two factors. First of all, the text samples are formed of $309$ tokens on average, being at least an order of magnitude longer than samples in typical dialectal corpora \cite{Ali-INTERSPEECH-2016, Samardzic-LREC-2016}. Second of all, the text samples can be discriminated in large part due to different word choices, as shown in the analysis of the most discriminative features provided in Section~\ref{sec_Experiments}. Word preference seems to become easily distinguishable when news samples of around $309$ tokens (multiple sentences) are used. In future work, we aim to determine if the same level of accuracy can be obtained when single sentences will be used as samples for training and testing.

\section*{Acknowledgments}
We thank reviewers for their useful comments. 

\bibliography{references}
\bibliographystyle{acl_natbib}

\end{document}